# Attitude and In-orbit Residual Magnetic Moment Estimation of Small Satellites Using only Magnetometer

Raunak Srivastava, Roshan Sah, Kaushik Das
TCS Research
Bangalore, India; +918292346729
srivastava.raunak@tcs.com

## ABSTRACT

Attitude estimation or determination is a fundamental task for satellites to remain effectively operational. This task is furthermore complicated on small satellites by the limited space and computational power available on-board. This, coupled with a usually low budget, restricts small satellites from using high precision sensors for its especially important task of attitude estimation. On top of this, small satellites, on account of their size and weight, are comparatively more sensitive to environmental or orbital disturbances as compared to their larger counterparts. Magnetic disturbance forms the major contributor to orbital disturbances on small satellites in Lower Earth Orbits (LEO). This magnetic disturbance depends on the Residual Magnetic Moment (RMM) of the satellite itself, which for higher accuracy should be determined in real-time. This paper presents a method for in-orbit estimation of the satellite magnetic dipole using a Random Walk Model in order to circumnavigate the inaccuracy arising due to unknown orbital magnetic disturbances. It is also ensured that the dipole as well as attitude estimation of the satellite is done using only a magnetometer as the sensor.

## INRODUCTION

Attitude estimation and determination is one of the most pivotal requirement of any satellite mission. Attitude, along with orbit determination determine the location of the satellite in space and its orientation in the orbit. However, attitude estimation of small satellites is complicated by their small size and small payload capacity. Small size brings in the additional constraint of limited computational resources, which are required to help determine the satellite's attitude. As a result, many SmallSat users are constrained to use a single sensor for attitude estimation purposes. Magnetometers are one of the most basic and reliable sources of information for measuring the attitude of satellites. They are widely used today due to their low cost and high reliability [1].

A lot of work has gone into estimation of satellite attitude through Kalman filter based methods using magnetometer and gyroscope as the sensors. Lefferts et al. [2] were one of the early researchers to estimate the satellite attitude using an Extended Kalman Filter (EKF). They examined several Kalman filter based schemes, which differed on the basis of their treatment of attitude error. Carletta et al. [3] proposed a method to estimate the attitude matrix through an unsymmetrical TRIAD algorithm using the measurements from a three-axis magnetometer. They arranged the complex matrix operations into a form of Faddeev algorithm that is then implemented on the FPGA core of the on-board computer. Habib [4] proposed an optimal fusion algorithm that optimally combines the output of different standard estimation algorithms. The computed output is stated to have a lower standard deviation than the previous algorithms. Jianqi et al. [5] proposed an Extended Kalman Filter (EKF) based orbit and attitude estimation algorithm using a magnetometer and a gyroscope. Filipski et al. [6] also used an EKF based on body-referenced representation of the state vector to estimate the satellite's attitude using only the geomagnetic field data. The algorithm's performance was evaluated through Monte Carlo simulations for a LEO nadir pointing satellite.

There are numerous examples of actual deployments of magnetometer and gyroscope based attitude determination systems on-board small satellites. Shou [7] described the development of an Attitude Determination and Control System (ADCS) for a microsatellite and verified its functioning using a Software in the Loop (SIL) method. The paper described the Kalman filter based ADCS functioning for three modes of the satellite- initialization mode, detumbling mode and mission mode. Similarly, Fritz et al. [8] described the ADCS design for the Cyclone Global Navigation Satellite System spacecraft. The detailed analysis comprised of software and hardware components of attitude estimation inline with the mission requirements was presented. A similar work by Lafontaine et al. [9] details the attitude determination system on-board European Space Agency's Proba spacecraft.

Although gyroscope as stated has been widely used on small satellites, it has its share of disadvantages. The gyro drift rate that leads to a fixed bias is a big problem for gyroscopes. It is also more prone to failures than a magnetometer and hence often requires a robust estimation technique. Adnane et al. [10] presented a Fault Tolerant Extended Kalman Filter (FTEKF) to improve the estimation reliability in the event of a sensor fault. A model of sensitivity factor was applied in the fault detection design of the filter to specifically detect gyroscopic failures. Soken et al. [11] also presented a modified EKF and Unscented Kalman Filter (UKF) as a robust estimation scheme against measurement faults. Zeng et al. [12] presented a Robust Adaptive Kalman filter using magnetometer and gyro measurements. A sequential scheme of orbit and attitude determination was proposed to reduce the impact of orbital errors on attitude estimation.

Apart from the satellite constraints, Low Earth Orbits (LEO) also pose several demanding challenges in the form of environmental disturbances, causing disruptions in attitude estimation and control. The largest of such disturbances in lower earth orbits are the magnetic disturbances caused due to the earth's magnetic field. This torque acts on the satellite due to its own Residual Magnetic Moment (RMM) or dipoles in the satellite body. It is important to consider this perturbation for accurate real-time estimation of the satellite attitude. Rawashdeh [13] developed a simulation tool named 'Smart Nanosatellite Attitude Propagator (SNAP)' to exhibit various environmental torques acting on a small satellite in the orbital environment. It was found that the magnetic disturbance on small satellites in low earth orbits is one order of magnitude higher than the remaining major disturbances (gravity gradient, atmospheric and solar).

Soken et al. [14] estimated the satellite's attitude, angular velocity and unknown constant components of external disturbance on a picosatellite using an Unscented Kalman Filter (UKF) by obtaining measurements from a magnetometer and a gyroscope. Gravity gradient torque, RMM, sun radiation pressure and aerodynamic drag were included in the unknown external disturbance. Ulrich et al. [15] presented Kalman based and nonlinear disturbance observer based techniques to estimate environmental perturbation torques acting on the satellite. The proposed estimation strategy was then applied to a gyroless Earth orbiting satellite mission of Proba-2. Soken et al. [16], [17] proposed a method for in-orbit estimation of time varying RMM of the satellite. They used a piecewise constant function to approximate the RMM and by using a simple approach, adapted the covariance matrix of Kalman filter to get better tracking accuracy in case of an unexpected abrupt change in the RMM without any adverse effect on the estimation accuracy.

The motivation behind this work lies in the fact that magnetic disturbances should be modelled and considered as an external torque in order to increase the accuracy of attitude estimation. However, most of the current works consider everything under one common umbrella of white Gaussian process noise. As magnetic perturbation arising due to the satellite's own RMM is paramount in LEO, modelling it should be a top priority. However, most of the research on this subject considers the satellite RMM to be a constant predetermined parameter and then models the magnetic disturbance. Accordingly, estimation algorithms either ignore the magnetic disturbance or consider it as a function of constant RMM. Since the RMM essentially changes slightly over time, it is better to estimate it in real-time instead of considering it as a constant or a piecewise constant parameter and then model the magnetic perturbation.

The main contribution of this work is thus to model the satellite RMM using a Random Walk model instead of considering it as a constant or piecewise constant function. The magnetic disturbance due to this varying RMM is then modelled and taken as an external torque while solving for the system dynamics. This paper proposes a body referenced EKF to estimate the satellite's RMM in real-time along with the satellite's attitude and angular velocity. The integration of magnetic disturbance estimation with body referenced EKF can be understood from the below given flowchart in Fig. 1. $B_{meas}$ represents the magnetic field measured by the magnetometer sensor while $B_{earth}$ represents the Earth's reference magnetic field. $m$ represents the satellite RMM while $q, \omega$ represent its attitude and angular velocity respectively. We believe that this approach will result in a more accurate attitude estimation with real-time estimate of the external torque on the satellite, as is evident from the results.

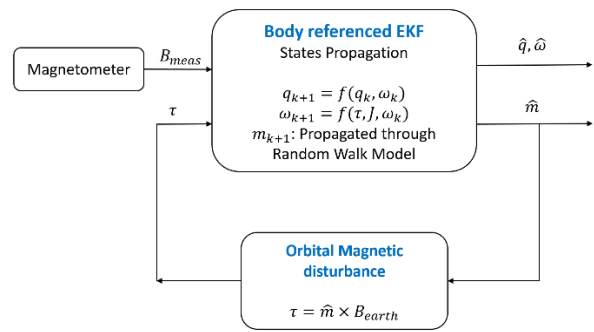

**Figure 1: Flowchart**

The organization of the paper is as follows. Section 'Satellite Model' details the kinematic and dynamic model of satellite attitude in an orbit. Also, the magnetic disturbance arising due to satellite Residual Magnetic Moment is presented. Section 'Attitude Estimation' presents the EKF used for predicting satellite attitude and angular velocity from a magnetometer. The limitations of the EKF in its current form is presented and a body referenced EKF to solve the problem is shown. Section 'Residual Magnetic Moment Estimation' presents the main contribution of this work. The problem of online RMM estimation is addressed and it is modelled as a random walk model. The resulting augmented EKF is presented for collective estimation of satellite attitude, angular velocity, and RMM. This results in an exact online measurement of non-stationary non-Gaussian magnetic disturbance. The simulation setup used for the study is explained in Section 'Simulation Steup' while the results are then shown in Section 'Results', followed by possible conclusions in Section 'Conclusion'.

**SATELLITE MODEL**

The Orientation of a satellite in its orbit depends on its attitude and angular velocity. We represent the satellite attitude in terms of quaternions instead of roll, pitch, and yaw angles in order to avoid singularities. This section gives a brief description of the quaternion used to represent satellite attitude, the kinematic and dynamic expressions for satellite model and lastly, the model considered for representing the magnetic disturbance acting on the satellite due to its own residual magnetic moment.

*Quaternion Representation of Attitude*

In the present study, the attitude is represented by the quaternion defined as

$$\bar{q} = \begin{bmatrix} \vec{q} \\ q_4 \end{bmatrix} = \begin{bmatrix} q_1 \\ q_2 \\ q_3 \\ q_4 \end{bmatrix}$$

Where

$$\vec{q} = \hat{n} \sin\left(\frac{\theta}{2}\right)$$

$$q_4 = \cos\left(\frac{\theta}{2}\right)$$

$\hat{n}$ represents the axis of rotation while $\theta$ represents the angle of rotation around the mentioned axis. Since the degrees of freedom remain 3, an additional constraint needs to be imposed on the quaternion to maintain the same. Accordingly,

$$\bar{q}^T \bar{q} = 1$$

With the quaternions in place, the attitude rotation matrix in terms of quaternions can be writeen as

$$A(q) = (|q_4|^2 - |\vec{q}|^2)I_3 + 2\vec{q}^T\vec{q} - 2q_4[\vec{q}] \qquad (1)$$

Where

$$[\vec{q}] = \begin{bmatrix} q_4 & q_3 & -q_2 & q_1 \\ -q_3 & q_4 & q_1 & q_2 \\ q_2 & -q_1 & q_4 & q_3 \\ -q_1 & -q_2 & -q_3 & q_4 \end{bmatrix}$$

*Satellite Equations of Motion*

Satellite attitude is represented by it's orientation with respect to the orbital frame. The rate of change of satellite attitude defines the angular velocity vector $\vec{\omega}$ and can be expressed as a quaternion product

$$\frac{d}{dt}\bar{q}(t) = \frac{1}{2}\bar{\omega} \otimes \bar{q}(t) = \frac{1}{2}[\bar{\omega}]\bar{q}(t) \qquad (2)$$

Where

$$\bar{\omega} = \begin{bmatrix} \vec{\omega} \\ 0 \end{bmatrix}$$

$$[\bar{\omega}] = \begin{bmatrix} 0 & \omega_3 & -\omega_2 & \omega_1 \\ -\omega_3 & 0 & \omega_1 & \omega_2 \\ \omega_2 & -\omega_1 & 0 & \omega_3 \\ -\omega_1 & -\omega_2 & -\omega_3 & 0 \end{bmatrix}$$

where $\bar{q}$ is the satellite attitude quaternion while $\vec{\omega}$ is the satellite body angular velocity with respect to the orbital frame.

Satellite attitude dynamics can be obtained from the Euler's law expressed in the body frame as

$$\dot{\omega} = J^{-1}(\tau - \omega^\times J\omega) \qquad (3)$$

Where

$$\omega^\times = \begin{bmatrix} 0 & -\omega_3 & \omega_2 \\ \omega_3 & 0 & -\omega_1 \\ -\omega_2 & \omega_1 & 0 \end{bmatrix}$$

$\tau$ is the net applied moment about the satellite centre of mass in the body frame, $J$ is the inertia matrix about the mass centre and $\omega$ is the body angular velocity with respect to inertial frame. Angular velocity of satellite relative to inertial frame can be expressed in the body frame as

$$\omega_{ib}^b = \omega_{io}^b + \omega_{ob}^b$$

$$\omega_{ib}^b = R_o^b \omega_{io}^o + \omega_{ob}^b$$

Where $\omega_{ib}^b$ is the satellite angular rate vector in the inertial frame while $\omega_{ob}^b$ is the satellite angular rate vector in the orbital frame. $R_o^b$ represents the rotation matrix from the orbital frame to the body frame and is obtained in terms of quaternion $\bar{q}$. $\omega_{io}^b$ is the orbital velocity in the inertial frame which remains constant in case of a circular orbit. It can be expressed as

$$\omega_{io}^o = \begin{bmatrix} 0 \\ -\omega_0 \\ 0 \end{bmatrix} = \begin{bmatrix} 0 \\ -\sqrt{\dfrac{GM}{R^3}} \\ 0 \end{bmatrix}$$

*Magnetic Disturbance*

Residual magnetic dipoles exist in the satellite structure as well as in the on board electronic equipment. Just like a magnetic dipole experiences a magnetic moment that seeks to align the dipole with the magnetic field lines, a magnetic moment also acts on the satellite due to its own residual magnetic dipole in the presence of Earth's magnetic field. This is the reason that this magnetic moment in the form of disturbance is strongest near the Earth's surface due to relatively strong magnetic field. Although passive control techniques make use of magnetic moment acting due to the Earth's magnetic field, it is essential that the exact nature and magnitude of such moments be known from before so that it no longer remains an unknown disturbance. This moment or torque acting on the satellite can be modelled as

$$M_m = m \times B_{Earth} \tag{4}$$

where, $m$ is the Residual Magnetic dipole of the satellite and $M_m$ is the torque acting on the satellite due to this dipole. Most of the previous works on satellite attitude estimation have either ignored the effect of this magnetic disturbance or have considered the satellite RMM to be constant while computing this magnetic disturbance. This leads to unrealistic computation of the magnetic disturbance, which is detrimental to accurate real-time estimation of the satellite attitude.

The current section gave an overview of the preliminaries required in order to estimate the satellite attitude in orbit and in order to solve the dynamic equations. With the basics ready, the next section details the EKF that is used for attitude estimation using a magnetometer as the sole sensor.

**ATTITUDE ESTIMATION**

This section presents the Extended Kalman Filter used for attitude and angular velocity estimation using a magnetometer. Initially an EKF based on standard state equations is presented which is later modified to cater to the problems arising due to an extra state in the EKF. The modified body referenced EKF is also explained.

*State Equation*

The satellite state can be considered to be composed of the attitude quaternion $\bar{q}$ and the angular velocity $\omega$.

$$x(t) = \begin{bmatrix} \bar{q} \\ \omega \end{bmatrix} \tag{5}$$

where, the attitude quaternion and angular velocity rates are given by Eq. 2 and 5 respectively. The nonlinear system requires an EKF for state estimation and considers a white Gaussian noise $w(t)$ as the process noise. This noise takes care of the imperfections in the dynamic modelling and acts as a substitute for any unknown disturbance which is Gaussian in nature.

$$\frac{d}{dt}x(t) = f(x(t), t) + w(t)$$

Such that

$$E(w(t)) = 0$$

$$E[w(t)w^T(t)] = Q(t)$$

where $Q(t)$ is the process noise covariance. The jacobian for state propagation using the above mentioned states can be expressed as

$$\phi(t) = \frac{\delta f}{\delta x} \Big|_{x = \hat{x}}$$

Where

$$\phi(t) = \begin{bmatrix} \dfrac{1}{2}\Omega(\omega) & \dfrac{1}{2}\Xi(\hat{\bar{q}}) \\ 0_{3\times 4} & J^{-1}((J\omega)^\times - \omega^\times J) \end{bmatrix}$$

$$\Xi(\hat{\bar{q}}) = \begin{bmatrix} q_4 & -q_3 & q_2 \\ q_3 & q_4 & -q_1 \\ -q_2 & q_1 & q_4 \\ -q_1 & -q_2 & -q_3 \end{bmatrix}$$

However, it happens that the covariance matrix for the above mentioned seven dimensional state vector is singular. This follows from the standard quaternion norm of

$$\bar{q}^T \bar{q} = 1$$

Thus, the covariance matrix can develop a zero or even a negative eigen value. One way to avoid such a situation is to use a state vector of a lower dimension. It is for this purpose that a body referenced EKF is used for the purpose of satellite attitude estimation.

*Body Referenced EKF*

Body referenced representation offers an alternative representation to the state vector and covariance matrix representation. The quaternion error in this representation is expressed not as the arithmetic difference between the actual and estimated quaternion, but as the small rotation required in the estimated quaternion to be equivalent to the true quaternion. Since this rotation is very small, the quaternions's fourth scalar component ($q_4$) is almost equal to unity. As a result, all the attitude information is available through the vector component ($\vec{q}$) of the quaternion. Thus, the seven-dimensional state vector reduces to six dimensional incremental representation of states in the form of state error. The error quaternion and error angular velocity is thus defined as

$$\delta q = \bar{q} \otimes \hat{\bar{q}}^{-1} \tag{6}$$

$$\Delta \omega = \omega - \hat{\omega} \tag{7}$$

The six dimensional state vector thus becomes

$$\tilde{x}(t) = \begin{bmatrix} \vec{\delta q} \\ \omega \end{bmatrix}$$

And the corresponding state perturbation for the body referenced representation is

$$\Delta \tilde{x} = \begin{bmatrix} \delta q \\ \Delta \omega \end{bmatrix}$$

where the first three terms refer to the vector component of the error quaternion whereas the last three terms refer to the angular velocity of the body in the inertial frame, expressed in the body frame. The error rate for quaternion can be expressed as

$$\delta \dot{q} = -\omega^\times \delta q + \frac{1}{2}\Delta \omega \tag{8}$$

Similarly, the error rate for angular velocity can be derived by differentiating Eq. 5 and ignoring higher order terms upon taking Taylor expansion. The error rate thus becomes

$$\Delta \dot{\omega} = J^{-1}[(J\omega)^\times - \omega^\times J]\Delta \omega \tag{9}$$

With these error rates as the state vector, the EKF takes the following form

$$\Delta \dot{\tilde{x}} = \phi \, \Delta \tilde{x} + w$$

where $\phi$ is the state transition matrix while $w$ is a white Gaussian noise. The state transition matrix has the following form

$$\phi(t) = \begin{bmatrix} -\omega^\times & \frac{1}{2}I_3 \\ 0_3 & J^{-1}((J\omega)^\times - \omega^\times J) \end{bmatrix}$$

*Available Measurement*

As stated before, the only sensor used for this study is a magnetometer. The measurement equation of a magnetometer can be given as

$$B_b = A(q)B_o = v$$

where $B_b$ represents the magnetic field at the body centre of mass in the body frame while $B_o$ represents the magnetic field at the body centre of mass in the orbital frame. While $B_o$ is calculated from the standard International Geomagnetic Refernce Model (IGRF), $B_b$ is the actual magnetometer reading installed on the satellite body. $v$ represents a white Gaussian noise inherent in the sensor. The rotation matrix from orbital to body frame can be computed from the satellite attitude quaternion as mentioned in Eq. 1.

The magnetic field measured by the magnetometer in the body frame ($B_b$) is compared with the estimate of magnetic field obtained through the rotation of standard magnetic field from orbital frame to the body frame ($A(q)B_o$). The difference, also called as innovation helps in the correction step of EKF to get the updated error rates of quaternion and angular velocity.

$$\Delta B = B_b - \hat{B}_b$$

For a small angle rotation, the attitude rotation matrix from Eq. 1 can be approximated as

$$A(q) = (|q_4|^2 - |\vec{q}|^2)I_3 + 2\,\vec{q}^T\vec{q} - 2q_4[\vec{q}]$$

$$A(\delta q) \approx I_3 - 2\delta q^\times$$

Thus, the innovation term can be written as

$$\Delta B = B_b - \hat{B}_b$$
$$= A(q)B_o + v - \hat{B}_b$$
$$= A(\delta q)A(q)B_o + v - A(q)B_o$$
$$= (I_3 - 2\delta q^\times)A(q)B_o + v$$
$$= 2(A(q)B_o)^\times \delta q + v$$

Thus, the measurement matrix becomes

$$C = [2(A(q)B_o)^\times \quad 0_3] \tag{10}$$

Thus, with the help of a single magnetometer as a sensor, it is possible to estimate the satellite attitude and angular velocity. However, as stated in the introduction, the mentioned EKF does not consider any perturbation or environmental disturbance of any kind. The next section details the method adopted to reduce the effect of magnetic disturbances by model the RMM using a Random Walk Model. Its integration into the body referenced EKF is also presented in the next section.

**RESIDUAL MAGNETIC MOMENT (RMM) ESTIMATION**

The biggest environmental disturbance on small satellites in Low earth orbits is due to the magnetic effect of the Earth. One of the prime reasons for this disturbance is the residual magnetic dipole of the satellite itself. However, this magnetic disturbance was erroneously considered a white Gaussian noise in the previously mentioned EKF.

This section seeks to model the magnetic torque acting on the satellite due to it's own RMM and then include it as an external torque while computing the satellite dynamics instead of considering it as a white Gaussian disturbance. However, it is essential to estimate the RMM of satellite in real time for modelling the magnetic disturbance as a known torque whose influence can then be easily circumvented. A Random Walk Model is used for the same purpose.

*Random Walk Model*

While propagating the states of an EKF, if the disturbance in the model is not stationary and Gaussian, it should be estimated and considered as an external disturbance while propagating the EKF. For this purpose, the unknown disturbance is augmented along with the remaining states and estimated. Accordingly, if the state equation is initially the standard equation

$$X(k+1) = F(X(k), U(k)) + D(k)$$

where $X(k)$ is the state vector, $U(k)$ the input vector and $D(k)$ is the stationary white Gaussian noise considered in the modelling process. In this case, the disturbance is unknown but its nature is assumed to be Gaussian. An EKF on such a propagation model does not give the desired results in case of non-Gaussian noise such as the magnetic perturbation given in Eq. 4. In such cases, the disturbance can be estimated along with the remaining states as shown below

$$\begin{bmatrix} X(k+1) \\ D(k+1) \end{bmatrix} = \begin{bmatrix} F(X(k), U(k), D(k)) \\ D(k+1) \end{bmatrix} + \begin{bmatrix} 0 \\ I \end{bmatrix} w_d(k) \quad (11)$$

The disturbance is modelled as a random walk model wherein the differential change in the disturbance is equal to the white Gaussian noise only. In such an implementation, the Gaussian noise is added only to the disturbance state instead of all the states as done previously. The disturbance in the remaining original state vector is now a function of the augmented state only.

*RMM estimation using EKF*

As stated above, magnetic disturbance should not be ignored and included as a white gaussan noise while propagating the satellite attitude and angular velocity error rates. Also, because the RMM of the satellite keeps changing over the period of orbit, it is better to estimate the RMM In real time than consider it as a constant for the entire duration. Once the RMM is estimated, the magnetic disturbance which is a function of RMM can be directly included as an external torque in the dynamics update of the satellite as in Eq. 5. Thus, the error in RMM is augmented in the state vector as

$$\Delta \tilde{x} = \begin{bmatrix} \delta q \\ \Delta \omega \\ \Delta m \end{bmatrix} \quad (12)$$

However, with the addition of error in RMM, the propagation model of the EKF changes. External torque in the form of Eq. 4 is now included in the dynamics propagation as shown in Eq. 3. Also, the white Gaussian noise is now only added in the disturbance state RMM instead of addition to all the remaining states. The new state transition looks like

$$\Delta \dot{\tilde{x}} = \phi \, \Delta \tilde{x} + \begin{bmatrix} 0_3 \\ 0_3 \\ I_3 \end{bmatrix} w$$

where $w$ is a white Gaussian noise. The modified differential equations for quaternion error rates and angular velocity error rates can be expressed as

$$\delta \dot{q} = -\omega^\times \delta q + \frac{1}{2} \Delta \omega \quad (13)$$

$$\Delta \dot{\omega} = J^{-1}[\Delta m \times B] + J^{-1}[(J\omega)^\times - \omega^\times J]\Delta \omega \quad (14)$$

RandomWalk model is applied to the satellite RMM in order to propagate it. However, the states of body referenced EKF are the differential change in the states and not the states themselves as shown in Eq. 12. Thus, the RMM and its differential (error in RMM) can be propagated as

$$\dot{m} = \eta$$

$$\Delta \dot{m} = 0$$

With all the three set of states (error in quaternion, error in angular velocity and error in RMM) in place, the new state transition matrix can be expressed as

$$\phi(t) = \begin{bmatrix} -\omega^\times & \frac{1}{2}I_3 & 0_3 \\ 0_3 & J^{-1}((J\omega)^\times - \omega^\times J) & -J^{-1}B^\times \\ 0_3 & 0_3 & 0_3 \end{bmatrix}$$

Since there is no new measurement added, the only change in measurement matrix is to accommodate the augmented state. Thus, the modified measurement matrix becomes

$$C = [2(A(q)B_o)^\times \quad 0_3 \quad 0_3] \qquad (15)$$

**SIMULATION SETUP**

Figure 2 describes the coordinate frames in use for the current simulation. The inertial frame is the Earth centered Inertial (ECI) frame with its origin at the earth's centre. The orbital frame is the North-East-Down (NED) frame at the satellite centre of mass. The satellite's body frame is such that the $z$ axis points towards the centre of Earth while $x$ axis points in the direction of it's velocity vector. $y$-axis completes the right-handed triad of the coordinate system. The satellite used for the current study is the M-SAT whose properties were used to generate the results. M-SAT has a dimension of $50cm \times 50cm \times 60cm$ and is hexagonal in shape. It has a mass of around $50kg$. M-SAT orbits in a LEO orbit at an altitude of around $400km$ from the Earth's surface with an orbit inclination of $40^o$.

For the purpose of attitude and in-orbit RMM estimation of the satellite, the satellite was simulated in the simulator STK in order to get its orbital positions and latitude, longitude, altitude details over a time period of $30min$. The latitude, longitude details give us the pinpoint location of the satellite in the ECI frame, which helps us get the Earth's magnetic field at that position according to the IGRF-12 model. This serves as the reference magnetic field which is used in the correction step of the EKF later on.

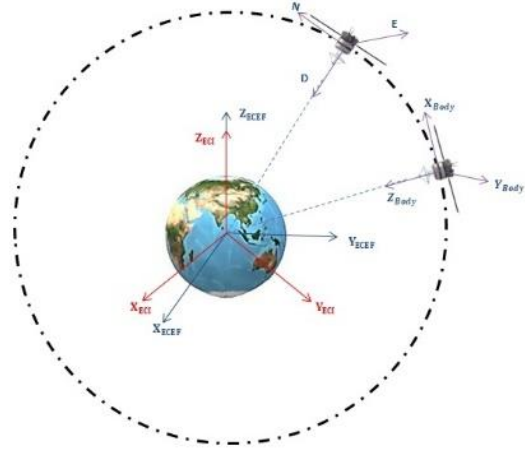

**Figure 2: Coordinate Frames**

**RESULTS**

This section puts forward the results obtained from the proposed method of RMM estimation. Firstly, quaternion and angular velocity estimates are shown when the RMM is considered fixed. A constant pre-determined value of RMM is taken and the external magnetic disturbance is considered. Thereafter, RMM is estimated in real time by propagating it using a RandomWalk model and the improvement in estimation results shown. The improved results depict the efficacy of RMM estimation using the proposed methodology.

*Without real-time RMM estimation*

Figure 3 and Fig. 4 show the attitude quaternion and angular velocity estimates of the satellite as obtained from the EKF. The simulation is run while considering the magnetic disturbance but with constant pre-determined value of the Residual magnetic moment of the satellite. The continuous lines depict the estimates while the dashed line depicts the actual ground truth unknown to the estimator. It can be seen that although the estimates try to follow the actual data, there are quite a few mismatches at various points in time. Moreover, the estimates have a lot of chattering due to the magnetic disturbance. We seek to rectify these shortcomings by attitude and angular velocity estimation while estimating the RMM also in real time.

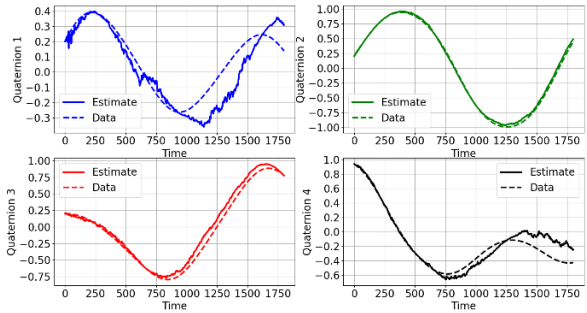

**Figure 3: Quaternion estimates with constant RMM**

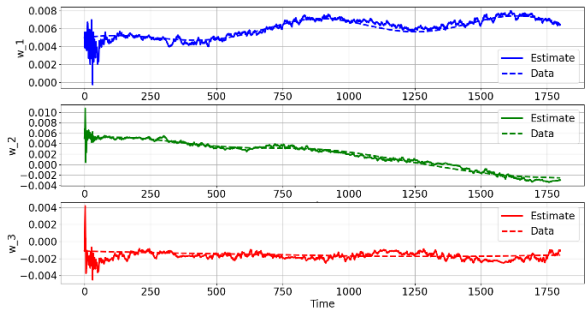

**Figure 4: Angular velocity estimates with constant RMM**

*With real-time RMM estimation*

This subsection presents the results of the method proposed in the current work. The EKF is augmented with the RMM components which are in turn propagated by a Random Walk model. The resulting estimates are shown.

Figure 5 shows the estimates of the four quaternions obtained from the EKF while considering the external magnetic disturbance. The continuous line shows the EKF estimate while the dashed line shows the actual modelled data which serves as the ground truth unknown to the estimator. It can be seen that the RMM augmented estimator is efficiently able to estimate the attitude quaternion despite the presence of magnetic disturbance. The first three quaternions represent the vector part of the attitude while the fourth quaternion represents the scalar part.

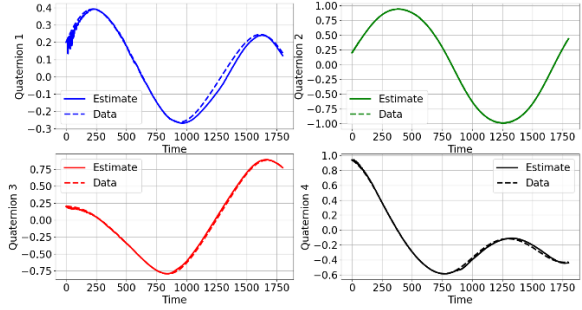

**Figure 5: Quaternion estimates**

Figure 6 shows the angular velocity estimates obtained from the EKF estimator. This velocity is in the body frame with respect to the inertial ECI frame. It can again be seen that the estimator is able to estimate the actual values of the angular velocities correctly. The continuous line shows the estimates while the dashed line shows the actual ground truth values of the angular velocity components unknown to the estimator. Initial chattering in the estimates is perhaps due to the large initial co-variance taken for the state estimates.

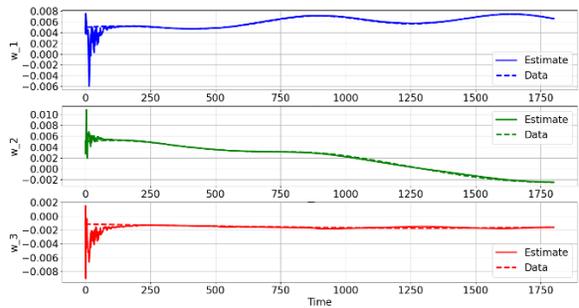

**Figure 6: Angular velocity estimates**

An important contribution of the current work is shown in the Fig. 7, which shows the satellite's residual magnetic moment estimates. Initial value of this RMM is shown as dashed lines which is usually considered as constant in most state of the art estimators. However, the current work uses a Random Walk model to propagate this RMM and hence model the external magnetic disturbance instead of considering it as a white Gaussian noise. The continuous line in Fig. 7 shows the estimate of RMM components obtained from the estimator. It can be seen that they tend to vary from the initial value of RMM and do not match exactly with the initial values. Thus, it is better to use real-time estimate of the time-varying RMM values instead of constant, pre-calibrated value of the same parameter. Reasonable estimates of the quaternion and angular velocities in Fig. 5 and Fig. 6 are testimonies to the accuracy and importance of this estimation.

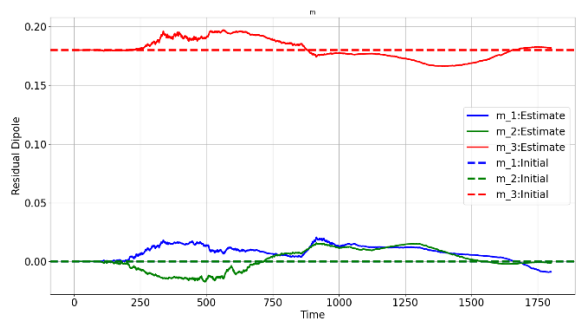

**Figure 7: RMM estimates**

Figure 8 shows the innovation components of the estimator. The dashed line shows the actual measurements which is of the order of $10^{-5}$ while the continuous line shows the innovation terms along the three component axes. The innovation is of the order of $10^{-6}$. This further solidifies the fact that the EKF estimator is correctly able to estimate the satellite attitude and in-orbit RMM in the presence of magnetic disturbance.

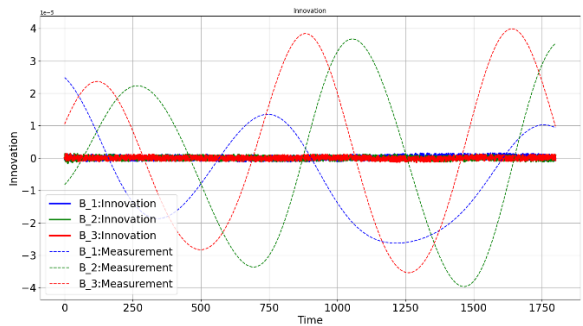

**Figure 8: Innovation**

**CONCLUSION**

This paper proposed a method to estimate the residual magnetic moment of the satellite in-orbit with the objective of accurately modelling the magnetic disturbance acting on the satellite. For the purpose of satellite attitude estimation, state of the art methods consider the RMM to be a constant precalibrated value while the RMM value tends to change during flight due to variations in the electric flow in the on board equipment. Because this variation is not very large and is completely random, it cannot be modelled in any way. This paper thus uses a Random Walk model to model the RMM and then uses the updated RMM value to accurately model the magnetic disturbance acting on the satellite in real-time.

The results of this exercise have been put forward. Accurate estimation of the satellite attitude and angular velocity show that the proposed method of in-orbit estimation of satellite RMM works well. The estimates of quaternion and angular velocity are much more accurate and smooth when the RMM is estimated real time as opposed to when it is considered as a constant. The estimate of RMM as shown above proves that the RMM indeed varies in flight, and it is hence more appropriate to estimate the RMM and therefore magnetic disturbance in flight for high accuracy of the satellite attitude estimation. Moreover, the entire estimation only uses a magnetometer as the sensor. Hence this is highly useful for small satellites which have limited space and computational power and hence support lesser sensors and actuators.

*References*

1.  J. Cote and J. de Lafontaine, "Magnetic-only orbit and attitude estimation using the square-root unscented kalman filter: application to the proba-2 spacecraft," in AIAA Guidance, Navigation and Control Conference and Exhibit, p. 6293, 2008.

2.  E. Leffens, F. L. Markley, and M. D. Shuster, "Kalman filtering for spacecraft attitude estimation," Journal of Guidance, Control, and Dynamics, vol. 5, no. 5, pp. 417–429, 1982.

3.  S. Carletta, P. Teofilatto, and M. S. Farissi, "A magnetometer-only attitude determination strategy for small satellites: Design of the algorithm and hardware in-the-loop testing," Aerospace, vol. 7, no. 1, p. 3, 2020.

4.  T. M. A. Habib, "A new optimal fusion algorithm for spacecraft attitude determination and estimation algorithms," The Egyptian Journal of Remote Sensing and Space Science, vol. 21, no. 3, pp. 305–309, 2018.

5.  W. Jianqi, C. Xibin, and S. Zhaowei, "Attitude and orbit determination for small satellite using magnetometer measurement," Aircraft Engineering and Aerospace Technology, 2003.

6.  M. Filipski and R. Varatharajoo, "Evaluation of a spacecraft attitude and rate estimation algorithm," Aircraft Engineering and Aerospace Technology, 2010.

7.  H. N. Shou, "Micro-satellite attitude determination and control subsystem design and implementation: Software-in-the-loop approach," in 2014 International Symposium on Computer, Consumer and Control, pp. 1283–1286, IEEE, 2014.

8.  M. Fritz, J. Shoer, L. Singh, T. Henderson, J. McGee, R. Rose, and C. Ruf, "Attitude determination and control system design for the

cygnss microsatellite," in 2015 IEEE Aerospace Conference, pp. 1–12, IEEE, 2015.

9. J. De Lafontaine, J. Buijs, P. Vuilleumier, P. Van den Braembussche, and K. Mellab, "Development of the proba attitude control and navigation software," in Spacecraft Guidance, Navigation and Control Systems, vol. 425, p. 427, 2000.

10. A. Adnane, H. Jiang, M. A. S. Mohammed, A. Bellar, and Z. A. Foitih, "Reliable kalman filtering for satellite attitude estimation under gyroscope partial failure," in 2018 2nd IEEE Advanced Information Management, Communicates, Electronic and Automation Control Conference (IMCEC), pp. 449–454, IEEE, 2018.

11. H. E. Soken, C. Hajiyev, and S.-i. Sakai, "Robust kalman filtering for small satellite attitude estimation in the presence of measurement faults," European Journal of Control, vol. 20, no. 2, pp. 64–72, 2014.

12. Z. Zeng, S. Zhang, Y. Xing, and X. Cao, "Robust adaptive filter for small satellite attitude estimation based on magnetometer and gyro," in Abstract and Applied Analysis, vol. 2014, Hindawi, 2014.

13. S. A. Rawashdeh, "Attitude analysis of small satellites using model-based simulation," International Journal of Aerospace Engineering, vol. 2019, 2019.

14. H. E. S̈oken and C. Hajiyev, "Estimation of picosatellite attitude dynamics and external torques via unscented kalman filter," Journal of Aerospace Technology and Management, vol. 6, no. 2, pp. 149–157, 2014.

15. S. Ulrich, J. Cˆot´e, and J. de Lafontaine, "In-flight attitude perturbation estimation for earth-orbiting spacecraft," The Journal of the Astronautical Sciences, vol. 57, no. 3, pp. 633–665, 2009.

16. H. E. Soken, S.-i. Sakai, and R. Wisniewski, "In-orbit estimation of time-varying residual magnetic moment," IEEE Transactions on Aerospace and Electronic Systems, vol. 50, no. 4, pp. 3126–3136, 2014.

17. H. E. Soken and S.-i. Sakai, "In-orbit estimation of time-varying residual magnetic moment for small satellite applications," in AIAA Guidance, Navigation, and Control (GNC) Conference, p. 4650, 2013.